\documentclass[letterpaper]{article} 
\usepackage{imghalluc}  
\usepackage{times}  
\usepackage{helvet}  
\usepackage{courier}  
\usepackage[hyphens]{url}  
\usepackage{graphicx} 
\usepackage{natbib}  
\usepackage{caption} 
\usepackage{algorithm}
\usepackage{algorithmic}

\usepackage{booktabs}
\usepackage{adjustbox}
\usepackage{multirow}
\usepackage{tabularx}
\usepackage{xspace}
\usepackage{arydshln}
\usepackage{amsmath}
\usepackage{placeins}
\usepackage{afterpage}
\usepackage{array}
\usepackage{amssymb}

\usepackage{xcolor}         
\usepackage{pifont}

\newcommand{\method}{I-HallA\xspace}
\newcommand{\benchmark}{I-HallA v1.0\xspace}

\usepackage{newfloat}
\usepackage{listings}
\DeclareCaptionStyle{ruled}{labelfont=normalfont,labelsep=colon,strut=off} 
\floatstyle{ruled}
\newfloat{listing}{tb}{lst}{}
\floatname{listing}{Listing}
\usepackage{hyperref} 
\nocopyright 

\title{Evaluating Image Hallucination in Text-to-Image Generation \\with Question-Answering
}
\author{
    Youngsun Lim\footnotemark[1],
    Hojun Choi\footnotemark[1],
    Hyunjung Shim
}
\affiliations{
    KAIST AI\\
    \{youngsun\_ai, hchoi256, kateshim\}@kaist.ac.kr
}

\begin{document}

\renewcommand{\thefootnote}{\fnsymbol{footnote}}
\maketitle
\footnotetext[1]{Equal contribution}

\begin{abstract}
Despite the impressive success of text-to-image (TTI) generation models, existing studies overlook the issue of whether these models accurately convey factual information. In this paper, we focus on the problem of image hallucination, where images created by generation models fail to faithfully depict factual content. To address this, we introduce I-HallA (Image Hallucination evaluation with Question Answering), a novel automated evaluation metric that measures the factuality of generated images through visual question answering (VQA). We also introduce I-HallA v1.0, a curated benchmark dataset for this purpose. As part of this process, we develop a pipeline that generates high-quality question-answer pairs using multiple GPT-4 Omni-based agents, with human judgments to ensure accuracy. Our evaluation protocols measure image hallucination by testing if images from existing TTI models can correctly respond to these questions. The I-HallA v1.0 dataset comprises 1.2K diverse image-text pairs across nine categories with 1,000 rigorously curated questions covering various compositional challenges. We evaluate five TTI models using I-HallA and reveal that these state-of-the-art models often fail to accurately convey factual information. Moreover, we validate the reliability of our metric by demonstrating a strong Spearman correlation ($\rho$=0.95) with human judgments. We believe our benchmark dataset and metric can serve as a foundation for developing factually accurate TTI generation models.
Additional resources can be found on our project page: \href{https://sgt-lim.github.io/I-HallA/}{https://sgt-lim.github.io/I-HallA}.

\end{abstract}

\section{Introduction}
As generative models \cite{gm:00, stablediff} continue to evolve, the demand for generating factual content alongside imaginary content has grown \cite{gm:03, reimagen}. In natural language generation, outputs with factual errors are classified as hallucinations, and considerable research has focused on mitigating this issue \cite{llmfaithfulness1, factscore}.

\begin{figure}[t]
\centering
\includegraphics[width=\columnwidth]{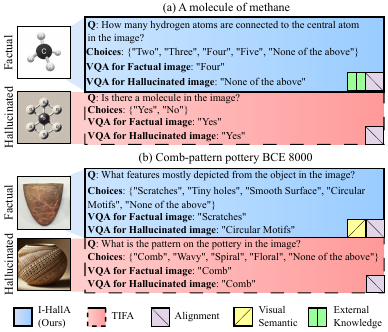}
\caption{Examples of the image hallucination and how I-HallA operates to evaluate it, along with a comparison to the existing metric, TIFA. \method can evaluate image hallucination by identifying factual information with two aspects: external knowledge and visual semantics. In contrast, TIFA hardly evaluates image hallucination as it relies solely on text prompts. \method assesses whether the VQA model can accurately answer questions about image hallucination. We use DallE-3 for the hallucinated images in this figure.}
\label{fig:main}
\end{figure}

\newsavebox{\imagebox}
\savebox{\imagebox}{\includegraphics[width=1.0\textwidth]{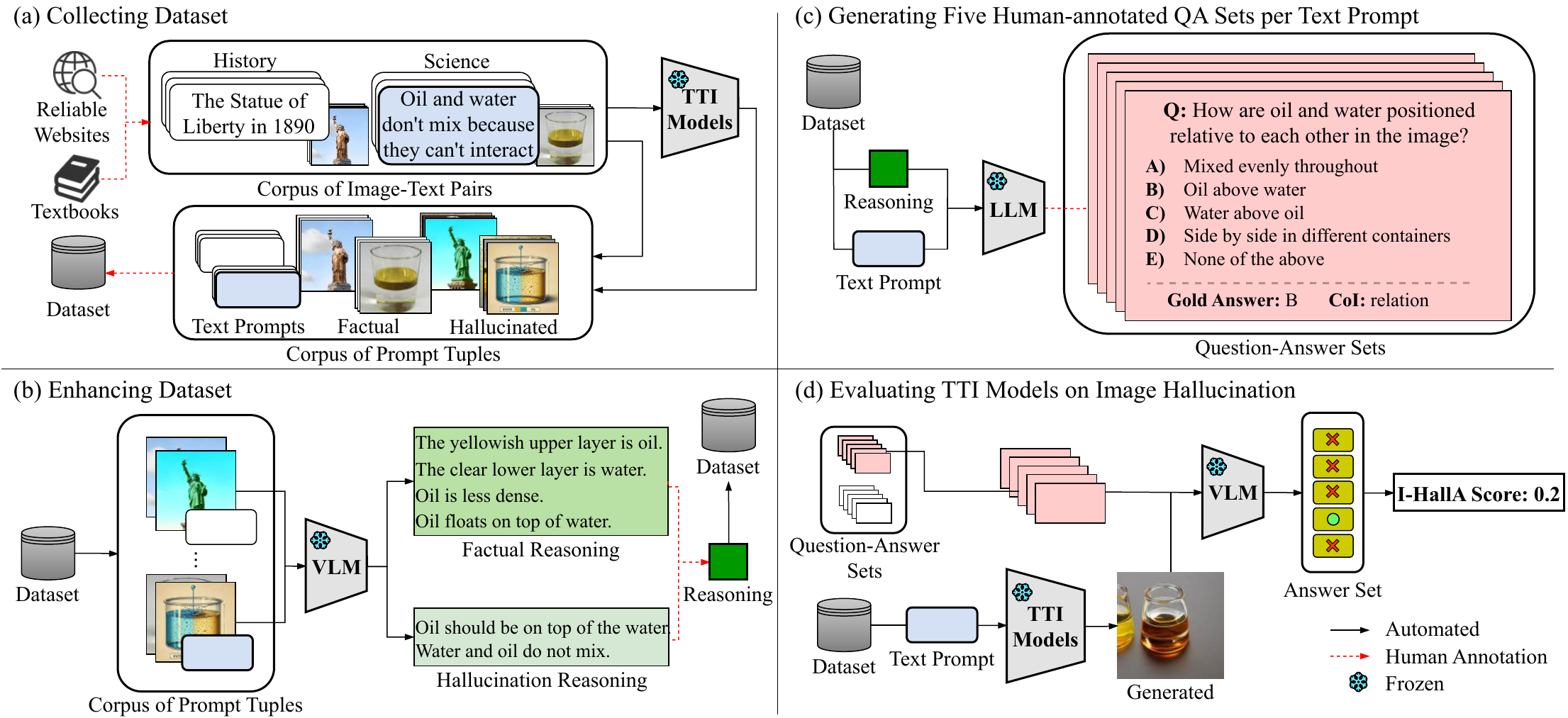}}
\begin{figure*}[t]
\centering
\usebox{\imagebox}
\caption{Overall pipeline of how I-HallA v1.0 is used for evaluating image hallucination: (a) Collect datasets containing prompts, factual images, and hallucinated images based on textbooks. (b) Enhance the collected dataset by leveraging the vast pre-trained knowledge and visual understanding capability of GPT-4o, adding reasoning about image hallucination to the datasets. (c) Input the prompt and reasoning into a language model to generate QA sets for evaluation. (d) Input the 5 QA sets per image and the target image into a vision-language model, and calculate the I-HallA score based on the number of correct answers. We employ GPT-4o for both the VLM and LLM.}
\label{main2}
\end{figure*}

Current text-to-image (TTI) models also struggle to accurately reflect factual information, generating incorrect images for given prompts, as illustrated in Figure \ref{fig:main}. This issue is becoming increasingly critical as TTI models are being actively utilized in industries and fields where factual accuracy is essential \cite{problem_nature}. For instance, if images against factual information are used in educational materials or the media, misinformation and misconceptions can spread rapidly, causing serious social side effects \cite{problem_google}.

While hallucinations have been primarily discussed in the language domain, relatively little research has addressed this issue in the context of image generation. 
This paper focuses on the unexplored issue of ``image hallucination,'' where generated images fail to reflect factual information \cite{imaghalluc}. To understand image hallucination in TTI models and guide future research directions, well-defined evaluation protocols and benchmark datasets are essential.
Recent studies have developed benchmarks and evaluation metrics to assess TTI models based on the alignment between text prompts and generated images \cite{vq2_seetrue}.
These evaluation protocols, such as TIFA \cite{TIFA}, focus only on the elements explicitly mentioned in the text prompt. 
However, verifying the factual alignment of generated images requires external knowledge beyond the prompt. As shown in Figure \ref{fig:main}-(a), the number of hydrogen atoms, though not mentioned in the prompt, is important factual information.
Additionally, current metrics struggle to distinguish between images that accurately represent factual information and those that simply match the text prompt, especially when polysemy introduces various interpretations.
As shown in Figure \ref{fig:main}-(b), while the generated image captures the idea of a ``comb pattern,'' it also includes several incorrect details, like small circles and decorations, which are not part of the factual image.

To address these limitations, we propose a three-stage pipeline to construct a new benchmark, I-HallA v1.0, with a new evaluation metric, I-HallA.
Unlike existing protocols, we leverage the vast knowledge base of GPT-4 Omni (GPT-4o) \cite{gpt4o} to assess factual information not mentioned in the prompt, such as the number of hydrogen and carbon atoms, molecular structure, and more. Additionally, we utilize the visual understanding capabilities of GPT-4o to discern factual visual semantics from potentially false ones, which is difficult to do with text prompts alone.

First, the dataset includes 200 prompts based on content from five science and history textbooks \cite{text_history1, text_history2, text_physics, text_biology, text_earthscience} to address factual information.
Textbooks, meticulously edited for educational purposes, represent years of accumulated knowledge and are among the most authoritative sources, making them a primary basis for this dataset.
Specifically, we use the textbooks' captions and corresponding figures as prompts representing factual information and factual images. This is because textbook figures are carefully curated, highly aligned with their captions, and thoroughly validated for factual accuracy.
The hallucinated images generated from these prompts in five TTI models \cite{stablediff, sd2.0, SDXL, dalle3} are compared against the factual images.
In total, we gather 1,200 images for all prompts, consisting of both factual and hallucinated images.

Secondly, we enhance the dataset by inputting each prompt and its corresponding image into GPT-4o to obtain factual information, referred to as ``reasoning,'' relevant to the prompt.
This process leverages GPT-4o's external knowledge beyond the prompt and considers visual semantics to distinguish details difficult to discern from text alone.
Lastly, we construct I-HallA, consisting of 1,000 multiple-choice question-answer (QA) sets to evaluate the extent of image hallucination in TTI models, using reasoning as a key input.
With GPT-4o as our VQA model, we input the generated image and corresponding questions for each prompt. The accuracy of the answers is then scored, with higher accuracy indicating fewer hallucinations.
We average QA scores across all prompts to evaluate TTI models on \benchmark.
In all three stages, a thorough human review validates the metric's legitimacy, though future use won't require it.

By applying our metric to various TTI models, we quantitatively measure the extent of image hallucination.
Experimental results show a strong correlation between our metric and human evaluation, with Spearman's $\rho$=0.95, indicating close alignment in assessing hallucination.
Our benchmark effectively addresses image hallucination, paving the way for further advancements in mitigating this issue.

\section{Related Works}
\label{sec:related}
\subsection{Hallucination in Language Generation} 
In language models, hallucination refers to the generation of unfaithful content to the given source material \cite{llmhallucsurvey}. As large language models (LLMs) increasingly produce text that closely resembles human writing, there has been a growing emphasis on developing benchmarks to evaluate and distinguish hallucinated content. For instance, FEVER \cite{fever} is a dataset used for fact-checking that utilizes Wikipedia as its knowledge source. HaluEval \cite{halueval} combines automated generation with human annotation to detect hallucinations.

Hallucination also occurs in large vision-language models (VLMs) such as LLaVA \cite{llava}, where visual features are input into LLMs to generate textual descriptions. In VLMs, hallucination refers to a mismatch between the factual details of images (e.g., object presence, attributes, spatial relations) and the corresponding generated text \cite{vlmsurvey}. Various studies evaluate this using metrics like BLEU \cite{bleu} or CIDEr \cite{cider}, or by querying VLMs about object presence \cite{pope}.
In contrast, we focus on evaluating hallucinations in image generation, where generated images fail to depict factual information accurately.
While hallucinations can occur in our pipeline during text generation, we mitigate this issue through rigorous human reviews. Further details are covered in Appendix B.

\subsection{Common Sense Reasoning in VLMs}
Some studies use benchmark datasets to assess whether VLMs possess commonsense knowledge when interpreting images. For instance, the WHOOPS \cite{whoops} and ROME \cite{rome} datasets are created by inputting intentional text prompts that defy common sense into TTI models, resulting in odd and unconventional images.

These studies differ from our focus, as their benchmarks evaluate the language generated by VLMs when interpreting images, rather than assessing TTI models.
Furthermore, by using counter-intuitive prompts to intentionally generate weird images, they do not address image hallucination, where TTI models fail to reflect factual information when given factually accurate prompts. Moreover, their concept of common sense differs from factual accuracy.
For example, one prompt in WHOOPS is ``A little girl standing in front of a blackboard with math formulas on it.'' While this scenario is factual, as a young child can solve math problems, WHOOPS argues that this prompt defies common sense. Thus, existing research on common sense relies on limited content to judge factual information, failing to fully address image hallucination.
More details are in Appendix C.

\subsection{Evaluating Text-to-Image Generation with Question Answering} 
CLIPScore \cite{clipscore} and DALL-Eval \cite{dalleval}, which are early studies measuring the alignment between generated images and text prompts to evaluate TTI models, commonly exhibit limitations due to the inherent constraints of CLIP (e.g., inability to count objects) or the restricted scope of evaluation criteria.

With the growing capabilities of foundation models, a new approach has developed that validates alignment in text-to-image generation by using VQA models on questions derived from the prompt.
For instance, TIFA \cite{TIFA} classifies elements of the text prompt into 12 categories and generates a set of questions and answers using GPT-3 \cite{gpt3}. These sets are used to evaluate the image by inputting both the image and questions into the VQA model like mPLUG \cite{mplug}.
VQ\textsuperscript{2} \cite{vq2_seetrue} extracts key information from the text prompt, generates related questions, and evaluates the text-image alignment by checking whether the image provides correct answers.
VPEval \cite{vpeval} enhances these alignment evaluation methods by incorporating object detection and optical character recognition, allowing for a more precise assessment.
Davidsonian Scene Graph \cite{dsg} breaks down the prompt into small propositions and represents the dependencies between these propositions in a graph, ensuring that the generated questions are not redundant.
However, existing benchmarks that focus solely on evaluating the alignment between text prompts and images often fail to detect external knowledge beyond the text and the factual visual semantics embedded in the image, which is required to address image hallucination.

\section{Methodology}
\label{sec:method}

\subsection{Image Hallucination}

Factual information refers to data that can be objectively verified and proven true based on evidence or reliable sources.
It is an important evaluation criterion in fields that require reliable and accurate information, such as education \cite{fact_important}. In this paper, we focus on image hallucination, a phenomenon where the images generated by TTI models fail to accurately reflect factual information.

Existing benchmarks that merely evaluate the alignment between the text prompt and the generated image are inadequate for properly assessing image hallucination. Their limitations in evaluating image hallucination can be summarized in two key points: 
\begin{itemize}
    \item The inability to evaluate factual information beyond the prompt.
    \item The difficulty in identifying accurate visual semantics.
\end{itemize}

As shown in Figure \ref{fig:main}, existing evaluation metrics, such as TIFA \cite{TIFA}, rely solely on text prompts, which limit their ability to consider factual information not explicitly stated in the prompt. In contrast, our metric leverages GPT-4o's capability to generate questions and answers based on external factual information that is not specified in the prompt but has already been trained into the model. 
This allows us to assess whether crucial factual information, though not mentioned in the prompt, is accurately reflected in the generated image.

Additionally, existing metrics cannot evaluate whether the visual semantics within an image are hallucinated.
This is because the polysemy of text prompts can generate images that are not visually factual while reflecting the word's meaning.
In such cases, metrics based solely on the text prompt fail to distinguish these visual semantics, making it impossible to assess image hallucination.
For instance, in the case of ``Comb-pattern pottery BCE 8000,'' the factual and hallucinated images of the comb-pattern are represented in Figure \ref{fig:main}-(b). However, the word ``comb-pattern'' corresponds to various visual designs and is not confined to a single pattern. Consequently, visual representations that do not match the specific form intended---potentially contradicting historical fact---can still be described as ``comb-pattern.'' This ambiguity complicates current evaluation metrics to assess the factual information.
More details are available in Appendix I.
In contrast, our pipeline utilizes the visual capabilities of GPT-4o by inputting both the prompt and image together to generate an evaluation metric based on factual information.
From factual images, we obtain the reasons why these images are considered factual, while from hallucinated images, we acquire discriminative information about how incorrect semantics differ from those in corresponding factual images, as illustrated in Figure \ref{main2}-(b).
This enables us to distinguish accurate visual semantics that reflect factual information among the many possible choices corresponding to the prompt.
Therefore, our approach allows for evaluating factual information, including visual semantics like patterns that cannot be fully conveyed through the text prompt alone.

\begin{table}[t]
    \centering
    \begin{adjustbox}{width=\columnwidth}
        \begin{tabular}{>{\centering\arraybackslash}m{2cm} >{\centering\arraybackslash}m{3cm} >{\centering\arraybackslash}m{2cm} ccc >{\centering\arraybackslash}m{1.5cm}}
            \specialrule{0.15em}{0pt}{1.5pt}
            \multirow{2}{*}{Domain} & \multirow{2}{*}{Category} & \multirow{2}{*}{Type} & \multicolumn{3}{c}{Level} & \multirow{2}{*}{Total} \\
            \cmidrule(lr){4-6}
             &  &  & Easy & Medium & Hard &  \\
            \specialrule{0.15em}{2pt}{2pt}
            \multirow{6}{*}{Science} & \multirow{2}{*}{Physics} & Factual & 26 & 3 & 4  & 33 \\
            \cdashline{3-7}
             &  & Hallucinated & 25 & 4 & 4  & 33 \\
            \cline{2-7}
             & \multirow{2}{*}{Biology} & Factual & 19 & 1 & 1  & 21 \\
            \cdashline{3-7}
             &  & Hallucinated & 17 & 2 & 2 & 21 \\
            \cline{2-7}
             & \multirow{1}{*}{Earth} & Factual & 40 & 2 & 4  & 46 \\
            \cdashline{3-7}
             & \multirow{1}{*}{Science} & Hallucinated & 33 & 4 & 9 & 46 \\ \specialrule{0.05em}{1.5pt}{1.5pt}
             & & - & 160 & 16 & 24 & - \\ 
              \specialrule{0.05em}{1.5pt}{0pt}
              \specialrule{0em}{0.8pt}{0.8pt}
              \specialrule{0.05em}{0pt}{1.5pt}
             & \multirow{1}{*}{Western \& Africa} & Factual & 14 & 1 & 0 & 15 \\
            \cdashline{3-7}
            \multirow{10}{*}{History} & \multirow{1}{*}{Ancient} & Hallucinated & 5 & 2 & 8 & 15 \\
            \cline{2-7}
             & \multirow{1}{*}{Western \& Africa} & Factual & 7 & 0 & 0 & 7 \\
            \cdashline{3-7}
             & \multirow{1}{*}{Medieval} & Hallucinated & 5 & 2 & 0 & 7 \\
            \cline{2-7}
             & \multirow{1}{*}{Western \& Africa} & Factual & 35 & 1 & 3 & 39 \\
            \cdashline{3-7}
             & \multirow{1}{*}{Modern} & Hallucinated & 24 & 2 & 13 & 39 \\
            \cline{2-7}
             & \multirow{1}{*}{Eastern} & Factual & 9 & 0 & 1 & 10 \\
            \cdashline{3-7}
             & \multirow{1}{*}{Ancient} & Hallucinated & 2 & 4 & 4 & 10 \\
            \cline{2-7}
             & \multirow{1}{*}{Eastern} & Factual & 16 & 0 & 0 & 16 \\
            \cdashline{3-7}
             & \multirow{1}{*}{Medieval} & Hallucinated & 4 & 3 & 9 & 16 \\
            \cline{2-7}
             & \multirow{1}{*}{Eastern} & Factual & 12 & 1 & 0 & 13 \\
            \cdashline{3-7}
             & \multirow{1}{*}{Modern} & Hallucinated & 10 & 1 & 2 & 13 \\  \specialrule{0.05em}{1.5pt}{1.5pt}
             & & - & 143 & 17 & 40 & - \\ \specialrule{0.15em}{1pt}{0pt}
        \end{tabular}
    \end{adjustbox}
    \caption{Statistical Analysis of I-HallA v1.0 Benchmark Dataset by Domain, Category, Type, and Difficulty Level. Each prompt's category is based on corresponding textbooks, broadly divided into science and history. The ``Type'' refers to whether each prompt's image is factual or hallucinated. The ``Difficulty Level'' is determined by making five type predictions for each prompt and its corresponding image using GPT-4o. Based on the number of correct predictions, the difficulty is categorized as follows: 0-1 correct predictions are classified as ``Hard,'' 2-3 correct predictions as ``Medium,'' and 4-5 correct predictions as ``Easy.''}
    \label{tab:t2}
\end{table}

\subsection{I-HallA v1.0: Benchmark for Evaluating Image Hallucination}

We propose a curated benchmark, I-HallA v1.0, and an evaluation metric, I-HallA, to assess image hallucination in TTI models.
To our knowledge, this is the first benchmark to evaluate image hallucination in TTI-generated images.

As a pioneering effort, our benchmark focuses on the educational domain, where the accuracy of factual information is crucial. Education serves as an ideal starting point for this benchmark study due to the broad and diverse use of factual data. Textbooks are specifically chosen as they encapsulate knowledge accumulated over time, providing well-structured and reliably categorized content. Within this domain, we choose science and history, two subjects that heavily rely on images for effective learning. As shown in Table~\ref{tab:t2}, in science, our benchmark covers Physics, Biology, and Earth Science. In history, it is organized by geographic regions---Eastern and Western \& African---and by periods, such as Ancient, Medieval, and Modern.

We introduce a three-stage pipeline to construct our benchmark dataset and evaluation metric. First, we collect a dataset to address the image hallucination in the educational domain. Next, we enhance this dataset by leveraging GPT-4o's pre-trained knowledge and visual understanding capabilities. Finally, based on the enhanced dataset, we develop a metric to evaluate image hallucination.

\begin{table}[t]
    \centering
    \begin{adjustbox}{width=\columnwidth}
        \begin{tabular}{ccccccccccc}
            \specialrule{0.15em}{0pt}{1.5pt}
            Domain & Color & Counting & Existence & Others & Posture & Relation & Scene & Shape & Size \\
            \specialrule{0.15em}{1.5pt}{2pt}
            Science & 30 & 49 & 106 & 16 & 11 & 91 & 100 & 73 & 24  \\ 
            \specialrule{0.05em}{1.5pt}{0.8pt}
            \specialrule{0.05em}{0.8pt}{1.5pt}
            History & 64 & 53 & 110 & 9 & 40 & 45 & 72 & 84 & 23 \\ 
            \specialrule{0.15em}{1.5pt}{0pt}        
        \end{tabular}
    \end{adjustbox}
    \caption{Statistical Analysis of the I-HallA Metric by Compositions of Interest (CoIs). The dataset includes a total of 1,000 question-answering sets with their corresponding CoI; 500 sets each for the science and history categories.}
    \label{tab:t3}
\end{table}

\subsubsection{Collecting Our Dataset}
The first stage for creating a benchmark is the collection process for initial datasets. 
We engage 10 graduate students to curate prompts from three science \cite{text_physics, text_biology, text_earthscience} and two history textbooks \cite{text_history1, text_history2}.
For each chapter in textbooks, participants extract up to 10 prompts and their corresponding factual images.
The prompt $P$ is either derived from textbook figures or selected based on unanimous agreement among participants that it represents key content in the chapter, such as important events, phenomena, or artworks.
If a figure corresponding to the prompt $P$ is in the textbook, it is collected as a factual image $I_f$. In rare cases where no such figure exists, we search for an image related to $P$ on verified websites, such as government-operated ones, and collect it as $I_f$.

Each participant inputs the curated prompt $P$ into five TTI models, generating 10 images per model.
All participants evaluate these images to identify image hallucination that contradicts factual information.
The image with the most pronounced hallucination, unanimously agreed upon by all 10 participants, is selected as the representative hallucinated image $I_h$ for that prompt and model, highlighting the most evident hallucination and revealing each model's limitations.

To conclude, our dataset consists of 200 tuples $\{$$P$$, $$I_f$$, $$I_{h_i}$$\}_{i=1}^{5}$, each containing a prompt $P$, a factual image $I_f$, and five hallucinated images $I_{h_i}$, where $i$ represents each TTI model. We collect 100 tuples from the science domain, and the remaining 100 are from history. In total, we assemble a set of 1,200 pairs, each consisting of a prompt and either a factual or hallucinated image.

\subsubsection{Enhancing Our Dataset}
The second stage involves enhancing the collected dataset using GPT-4o to evaluate factual information better.
Leveraging GPT-4o, pre-trained on vast data, and equipped with visual understanding, we develop a dataset incorporating external knowledge and visual semantics beyond the text prompt.
For each prompt $P$, we input two sets---($P$, $I_f$) and ($P$, $I_h$)---into GPT-4o to obtain three aspects: responses, reasonings, and difficulty levels. 
We use hallucinated images $I_h$ generated by DallE-3 \cite{dalle3} to capture hallucinations produced even by the latest TTI model.
To determine the difficulty levels, we performed the same process five times independently.

The response, determined by GPT-4o, categorizes the input image as either ``factual'' if accurate or ``hallucinated'' if not.
The reasoning provides the factual justification for this response, incorporating external knowledge and visual semantic information. Thus, for each prompt, two types of reasoning are provided: one for correctly identifying a ``factual'' image and another for identifying a ``hallucinated'' image.
Difficulty levels are determined by evaluating GPT-4o's accuracy across five inference attempts in discerning the factual information related to the given prompt and image.
Using the initial dataset labels as the ground truth, we compare GPT-4o's response to determine correctness. Prompts and images are classified as ``Hard'' if GPT-4o answered correctly 0-1 times, ``Medium'' if correct 2-3 times, and ``Easy'' if correct 4-5 times. 
We store the reasonings only when GPT-4o provides the correct response.

10 human annotators thoroughly review and refine all reasonings into a final, well-expressed version, retaining only the unanimously agreed-upon parts.
In cases where GPT-4o fails to provide any correct response, the reasoning is developed through group discussion and consensus among all participants. More details are available in Appendix B.
Consequently, our dataset consists of 200 tuples $\{$$P$$, $$I_f$$, $$I_{h_i}$$, $$R$$, $$D$$\}_{i=1}^{5}$, where the reasoning $R$ and difficulty levels $D$ have been added to the previously collected dataset.

\subsubsection{I-HallA: An Evaluation Metric Using Question-Answering}
The final stage involves developing the evaluation metric, I-HallA, to evaluate the factual accuracy of images generated by TTI models.
It employs five multiple-choice QA sets to assess image hallucination based on the curated dataset from previous stages.
To analyze the benchmark and results, we introduce classification criteria called Compositions of Interests (CoIs) to categorize the QA sets. We select the most relevant compositions from existing TTI evaluation studies \cite{TIFA, genai} that are closely related to image hallucination: \emph{color, counting, existence, others, posture, relation, scene, shape, and size}. 
The ``others'' category applies when a given QA set does not fit into the other CoIs.

To generate the QA sets, we input the prompt $P$, reasoning $R$, and CoIs into GPT-4o. Based on the reasoning, GPT-4o generates five multiple-choice QA sets per prompt. Each QA set consists of a question targeting factual information, five answer choices (with the fifth option being ``None of the above''), and the correct factual answer as the gold answer.
Simultaneously, GPT-4o generates the most relevant CoI for each QA set.
The QA set generation follows two key guidelines: 
First, the more factual information an image contains, the more correct answers it should provide, thereby yielding higher scores for more factual images.
Second, qualitative information that cannot be visually verified through the image should be excluded in the questions.
The 10 participants then review the generated QA sets to ensure that they adhere to these two guidelines. Any disagreements among participants lead to revisions. This process results in 1,000 QA sets and their corresponding CoIs across 200 prompts.

To calculate the score using I-HallA (I-HallA score), we input the image to evaluate, along with the question and five answer choices, into a VQA model and compare the model's response with the gold answer. There are five questions per image, and the I-HallA score ranges from 0 to 1.
The formula can be expressed as follows:
\begin{equation}
    \frac{1}{|\mathcal{P}|} \sum_{p \in \mathcal{P}} \left( \frac{1}{|Q^p|} \sum_{(q, c, g) \in (Q^p, C^p, G^p)} \mathbb{I}(\texttt{VQA}(p, q, c) = g) \right)
\end{equation}
$\mathbb{I}(\cdot)$ is an indicator function that returns 1 if the condition is satisfied.
\texttt{VQA} is the VQA model's prediction for the given QA.
$\mathcal{P}$ is a set of prompts.
For the given prompt $p$, $Q^p$, $C^p$, and $G^p$ represent the sets of questions, choices, and gold answers, respectively.
$|\cdot|$ is the number of elements in a set.

\begin{figure}[!t]
\centering
\includegraphics[width=1.0\columnwidth]{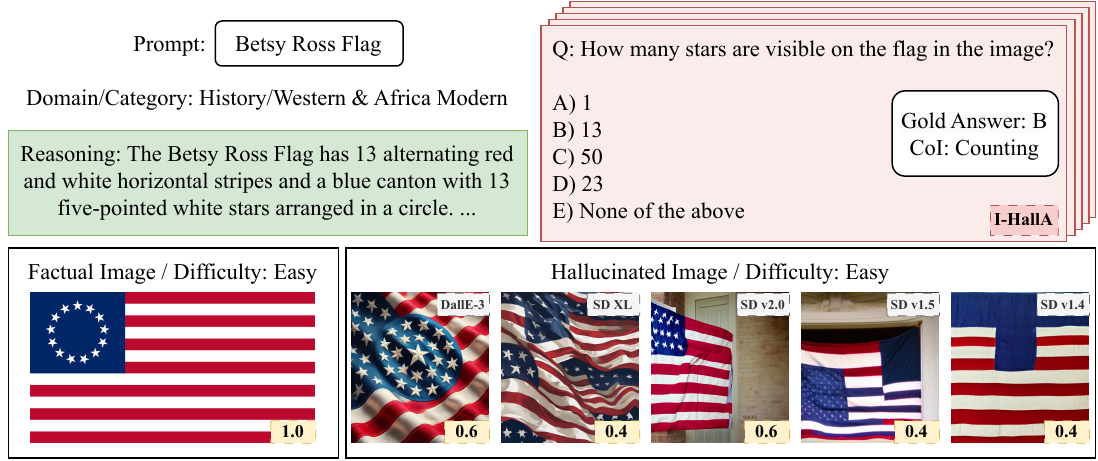}
\caption{Overview of I-HallA v1.0: The upper section presents the prompt, domain, category, reasoning, and I-HallA results for five QA sets. The lower section compares a factual image with hallucinated outputs from five TTI models, indicating difficulty levels and I-HallA scores. I-HallA scores shown in the bottom-right box of each image remain unchanged across the three trials.}
\label{fig:new_fig5}
\end{figure}

\section{Experiments}
\label{sec:exp}
In this section, we first analyze the statistical characteristics of \benchmark across various categories, difficulty levels, and a list of CoIs.
We evaluate the recent five text-to-image models with \method, emphasizing our metric robustly and accurately assesses image hallucination.
The models include DallE-3 \cite{dalle3}, Stable Diffusion v1.4, Stable Diffusion v1.5 \cite{stablediff}, Stable Diffusion v2.0 \cite{sd2.0}, and Stable Diffusion XL-base v1.0 \cite{SDXL}. 
Additionally, through human evaluation, we demonstrate that our method strongly correlates with human judgments on evaluating image hallucination.
For all experiments, we utilize GPT-4o as the VQA model for the \method. 


\begin{table}[t]
    \centering
    \begin{adjustbox}{width=\columnwidth}
        \begin{tabular}{ccccc}
            \specialrule{0.15em}{0pt}{1.5pt}
            \multirow{2}{*}{Models} & \multicolumn{2}{c}{I-HallA Score} & \multicolumn{2}{c}{I-HallA Score$^\dagger$} \\
            \cmidrule(lr){2-5}
            & Science & History & Science & History \\
            \specialrule{0.15em}{1.5pt}{2pt}
            SD v1.4 & 0.353 $\pm$ 0.002 & 0.535 $\pm$ 0.013 & 0.033 $\pm$ 0.012 & 0.110 $\pm$ 0.010 \\ 
            \specialrule{0em}{1.5pt}{0pt}\specialrule{0em}{0pt}{1.5pt}
            SD v1.5 & 0.309 $\pm$ 0.011 & 0.533 $\pm$ 0.004 & 0.030 $\pm$ 0.017 & 0.117 $\pm$ 0.021 \\
            \specialrule{0em}{1.5pt}{0pt}\specialrule{0em}{0pt}{1.5pt}
            SD v2.0 & 0.336 $\pm$ 0.006& 0.540 $\pm$ 0.014 & 0.027 $\pm$  0.021 & 0.120 $\pm$ 0.010\\
            \specialrule{0em}{1.5pt}{0pt}\specialrule{0em}{0pt}{1.5pt}
            SD XL & 0.398 $\pm$ 0.015 & 0.579 $\pm$ 0.012 & 0.077 $\pm$ 0.050 & 0.110 $\pm$ 0.066 \\
            \specialrule{0em}{1.5pt}{0pt}\specialrule{0em}{0pt}{1.5pt}
            DallE-3 & 0.661 $\pm$ 0.020 & 0.666 $\pm$ 0.003 & 0.227 $\pm$ 0.029 & 0.133 $\pm$ 0.031 \\
            \specialrule{0em}{1.5pt}{0.3pt}\cdashline{1-5}\specialrule{0em}{0.3pt}{1.5pt}
            Factual & \textbf{0.856} $\pm$ 0.002 & \textbf{0.873} $\pm$ 0.006 & \textbf{0.517} $\pm$ 0.038 & \textbf{0.533} $\pm$ 0.015 \\ 
            \specialrule{0.15em}{1.5pt}{0pt}        
        \end{tabular}
    \end{adjustbox}
    \caption{Our benchmark evaluation results on existing TTI models and factual images; We compute the I-HallA score by averaging ratio-based scores across 100 prompts per category (science and history). $\dagger$ means scoring as incorrect if even one out of five QA sets is wrong for each prompt. Each experiment is conducted three times.}
    \label{tab:t4}
\end{table}

\subsection{Benchmark Analysis} 
To illustrate the comprehensive scope of \benchmark, we provide an analysis spanning all categories, difficulty levels, and compositions. 
Additionally, we demonstrate that GPT-4o can be effectively used to develop \benchmark.
For more details on \benchmark, please refer to Appendix H.

\subsubsection{Statistics and diversity}


\begin{figure*}[t]
\centering
\includegraphics[width=1.0\textwidth]{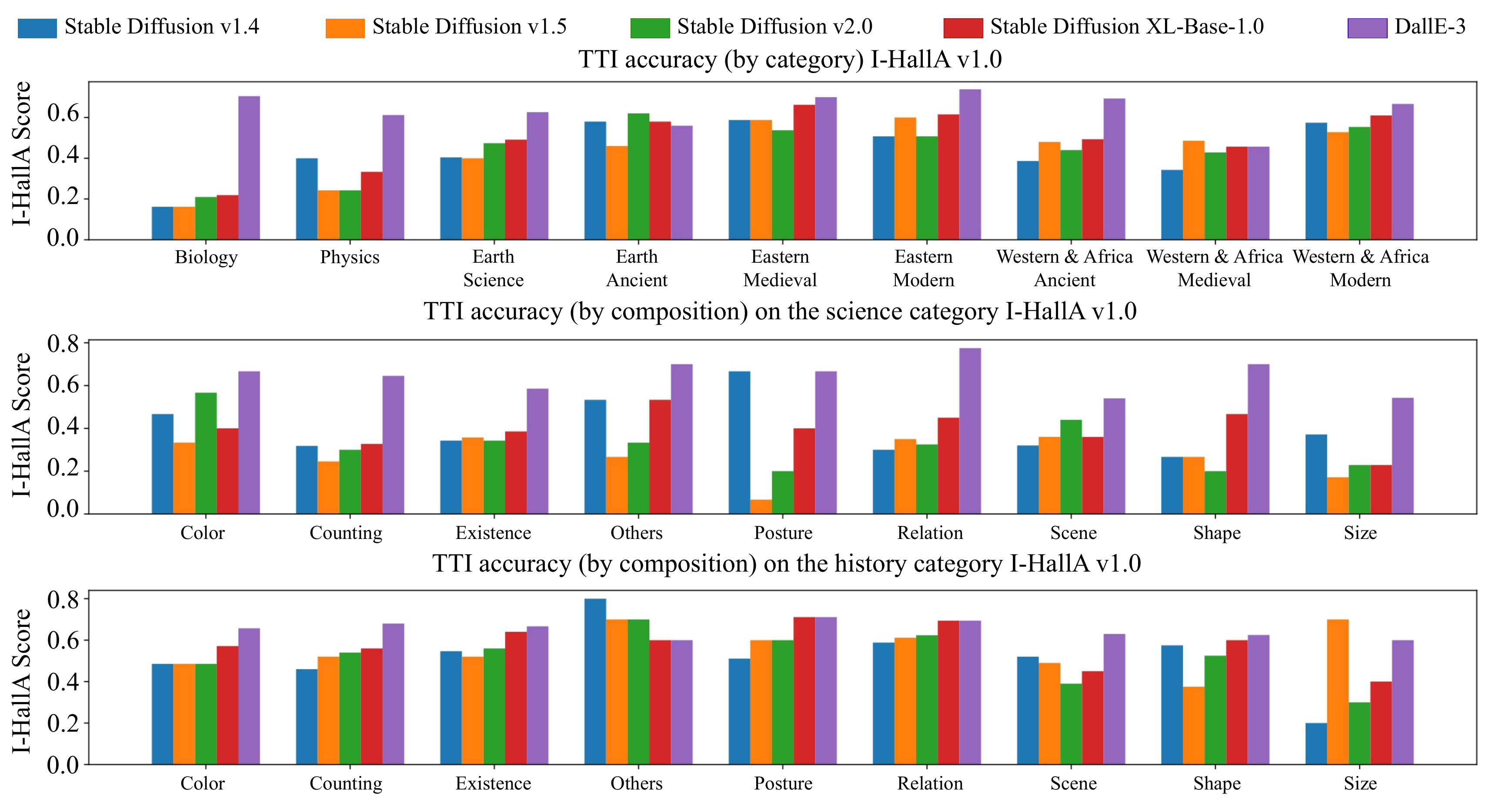}
\vspace{-0.15in}
\caption{I-HallA scores from five different TTI models across different categories and compositions. The factual information of images generated by TTI models using the prompts from I-HallA v1.0 is evaluated using the I-HallA metric. The I-HallA scores in this figure represent the average scores of each TTI model, calculated across different categories and compositions.}
\vspace{-0.1in}
\label{fig:fig3}
\end{figure*}

As shown in Table~\ref{tab:t2}, in the science domain, we collect 33, 21, and 46 prompts from physics, biology, and earth science textbooks, respectively. In the history domain, we collect prompts from two textbooks. Specifically, we gather 15, 7, 39, 10, 16, and 13 prompts from the Western \& African/Ancient, Western \& African/Medieval, Western \& African/Modern, Eastern/Ancient, Eastern/Medieval, and Eastern/Modern sections, respectively.
The number of images collected per prompt includes one factual image and five hallucinated images from different TTI models, totaling six images per prompt. Therefore, a total of 1,200 images are included in \benchmark.

Additionally, \method provides 1,000 questions with their corresponding compositions of interest, categorized into 9 types, as shown in Table~\ref{tab:t3}. In science, the occurrences are color (30), counting (49), existence (106), others (16), posture (11), relation (91), scene (100), shape (73), and size (24); while in history, they are color (64), counting (53), existence (110), others (9), posture (40), relation (45), scene (72), shape (84), and size (23).

\subsubsection{GPT-4o's ability on image hallucination}
In our study, we employ GPT-4o to generate reasoning and analyze the difficulty for I-HallA v1.0.
GPT-4o assesses whether an image is factual or hallucinated based on the prompt, with hallucinated images generated by the DALL-E 3.
Based on the three difficulty levels, GPT-4o classified 160 image-prompt pairs as Easy, 16 as Medium, 24 as Hard in the science domain; 143 as Easy, 17 as Medium, and 40 as Hard in history. When treating ``Hard'' pairs as incorrect, accuracy is 88\% in science and 80\% in history. If cases, where GPT-4o fails to answer any questions correctly, are treated as incorrect, accuracy increases to 91.5\% in science and 84.5\% in history. 

The higher number of ``Easy'' pairs suggests GPT-4o's strong ability to judge factual information, with ``Easy'' pairs being $\times4$ in science and $\times2.5$ in history compared to the total of others.
Moreover, as the reasoning and QA sets are thoroughly refined through human review, our benchmark 
is well-equipped to evaluate image hallucination.


\begin{figure}[!t]
\centering
\includegraphics[width=1.0\columnwidth]{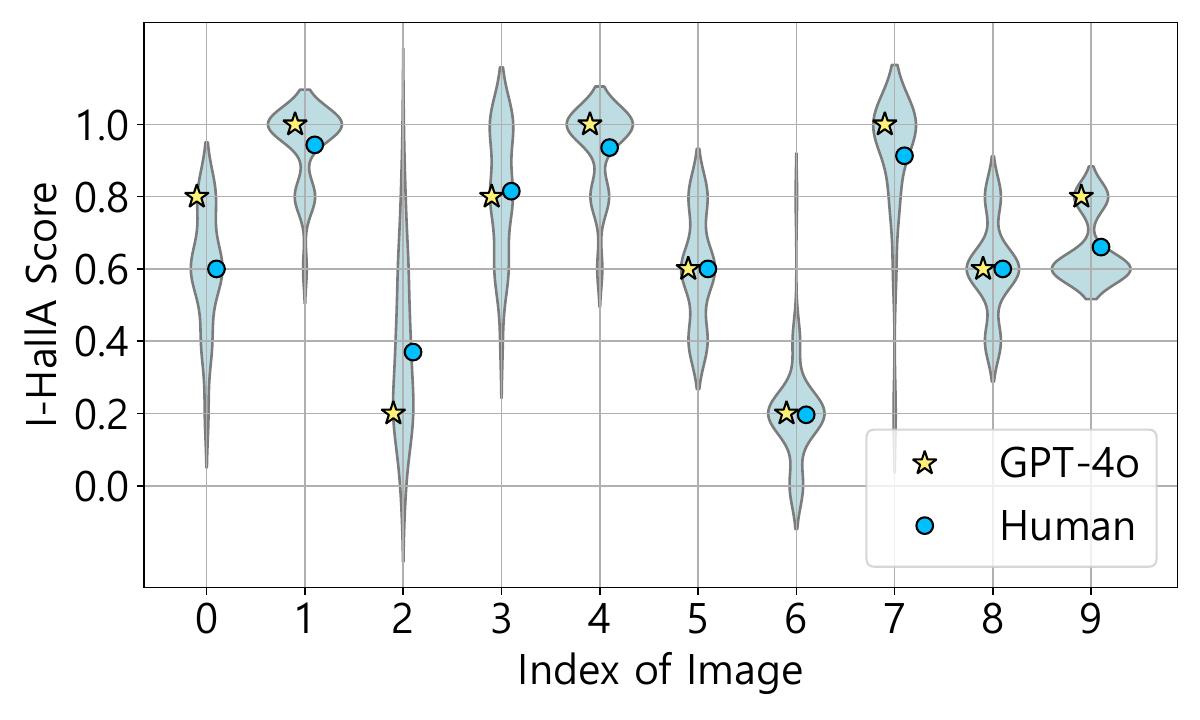}
\vspace{-0.25in}
\caption{Plot of I-HallA scores from GPT-4o and human evaluations.
Blue circles indicate average scores from 53 participants per question, with score distributions depicted via violin plots. Stars emphasize GPT-4o's results, which closely align with human judgments, demonstrating a strong correlation between the model and human evaluators.}
\label{fig4}
\end{figure}


\subsection{Evaluating Text-to-Image Models}
Using our I-HallA v1.0, we demonstrate that all five latest TTI models suffer from image hallucination.
By analyzing I-HallA scores, we quantitatively show how each model reflects factual information differently, proving that our benchmark objectively evaluates image hallucination. Furthermore, by categorizing the I-HallA scores across different categories and CoIs, we analyze specific situations where each model tends to produce image hallucinations.

Table~\ref{tab:t4} presents the average \method score and standard deviation from three trials for various TTI models on \benchmark. 
Higher scores indicate better performance in reflecting factual information without image hallucination.
DallE-3 outperforms the Stable Diffusion models in mitigating hallucination across all subjects. Even under the strict standard ($\dagger$), where one incorrect answer results in failure, DallE-3 remains the top performer.
I-HallA scores are generally higher in history than in science.
These scores allow us to assess how effectively current TTI models handle image hallucination.
Factual images score in the high 80s, much higher than the average of 0.411 in science and 0.570 in history for the five TTI models, demonstrating that our metric effectively measures factual information.
Still, they fall short of perfect, likely due to noise in I-HallA creation or VQA model limitations, which we aim to address in future work.


Figure \ref{fig:fig3} shows the impact of various TTI models on image hallucination across different categories and compositions.
The top graph displays average scores by category, while the middle and bottom graphs represent the average scores for each composition in science and history.
Models with larger parameters and newer architectures tend to have higher scores, with DallE-3 generally outperforming other models.
However, exceptions exist, such as in the ``Eastern Ancient'' category, where Stable Diffusion v2.0 scores higher than Stable Diffusion XL-base v1.0 or DallE-3.

In the science and history domains, the I-HallA score for ``Posture'' and ``Size'' compositions, respectively, is highest in the Stable Diffusion models, despite their generally lower image quality.
This suggests that these models effectively reflect factual information, as our metric evaluates factual accuracy rather than image quality.
Therefore, even TTI models that generate high-quality images can score lower if they fail to meet factual criteria.
For the ``Others'' composition in the history domain, the high \method scores of Stable Diffusion v1.4 might be influenced by a smaller sample size.

\vspace{-0.25in}
\subsection{Exploring the reliability of \method through human evaluation}
By calculating the correlation between the previously mentioned experimental results and the human evaluation of I-HallA v1.0, we demonstrate that our benchmark aligns well with human judgment in assessing image hallucination.
For more details on the user study, please refer to Appendix G.


We randomly sample 10 prompts from \benchmark, including 4 factual and 3 hallucinated images from each of two great TTI models in \method: DallE-3 and Stable Diffusion XL-base v1.0. 
We collect I-HallA scores from 53 participants, guiding them to answer questions based on the images provided, following the same approach as GPT-4o.
Figure~\ref{fig4} shows the average \method scores and standard deviations for each image. Even the image with the largest score difference shows a variance of only about 0.2, indicating that the results are very similar to human evaluations.

To verify this quantitatively, we calculate correlations between GPT-4o's I-HallA scores and human judgments on image hallucination. For the three metrics---Pearson's $r$, Spearman’s $\rho$, and Kendall’s $\tau$---we observe very high correlations of 0.952, 0.950, and 0.889, respectively.

\section{Conclusion}
\label{sec:conclusion}
In conclusion, we propose the I-HallA v1.0 benchmark as the first to address image hallucination in text-to-image generation by evaluating factual information. This benchmark overcomes the limitations of previous methods, which could not accurately assess factual information.
Developed through a three-stage pipeline using GPT-4o and thorough human review, it evaluates hallucination in 200 factual and 1,000 hallucinated images from five text-to-image models.
Our results confirm that the benchmark effectively measures image hallucination and aligns well with human judgment.
We hope that our benchmark and evaluation metric will be instrumental in resolving image hallucination in the future.

\section*{Acknowledgements}
This research was supported by 
Institute of Information \& communications Technology Planning \& Evaluation (IITP) grant funded by the Korea government (MSIT) (RS-2019-II190075 Artificial Intelligence Graduate School Program (KAIST), No.2021-0-02068 Artificial Intelligence Innovation Hub, No. RS-2024-00457882 National AI Research Lab Project), 
the Basic Science Research Program through the National Research Foundation of Korea (NRF) funded by the MSIP (NRF-2022R1A2C3011154),
IITP grant funded by the Korea government (MSIT) and KEIT grant funded by the Korea government (MOTIE) (No. 2022-0-00680, No. 2022-0-01045), RS-2023-00219019, and
Basic Science Research Program through the National Research Foundation of Korea (NRF) funded by the Ministry of Education (RS-2024-00394173).

\bibliography{imghalluc}

\appendix


%

\twocolumn[
    \begin{center}
        \LARGE{\bf{Supplementary Material for Evaluating Image Hallucination \\in Text-to-Image Generation with Question-Answering\\}}
    \end{center}
]

\begin{figure}[!ht]
    \centering
    \includegraphics[width=1.0\columnwidth, height=\textheight, keepaspectratio]{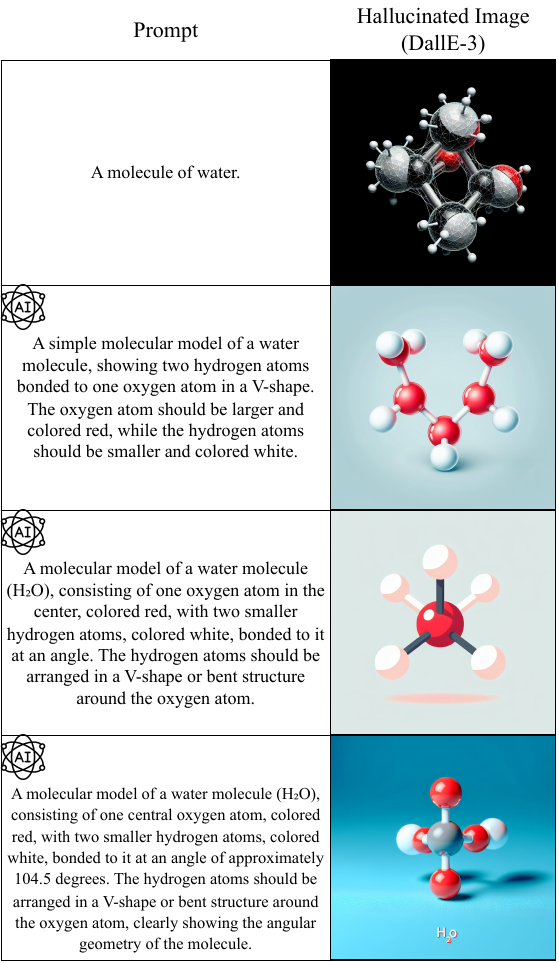} 
    \caption{The difficulty in resolving image hallucination through prompt refinement alone. No matter how detailed and lengthy the input text prompt is, current text-to-image models struggle to accurately reflect factual information.}
    \label{fig2}
\end{figure}

\begin{figure*}[!htb]
    \centering
    \includegraphics[width=1.0\textwidth]{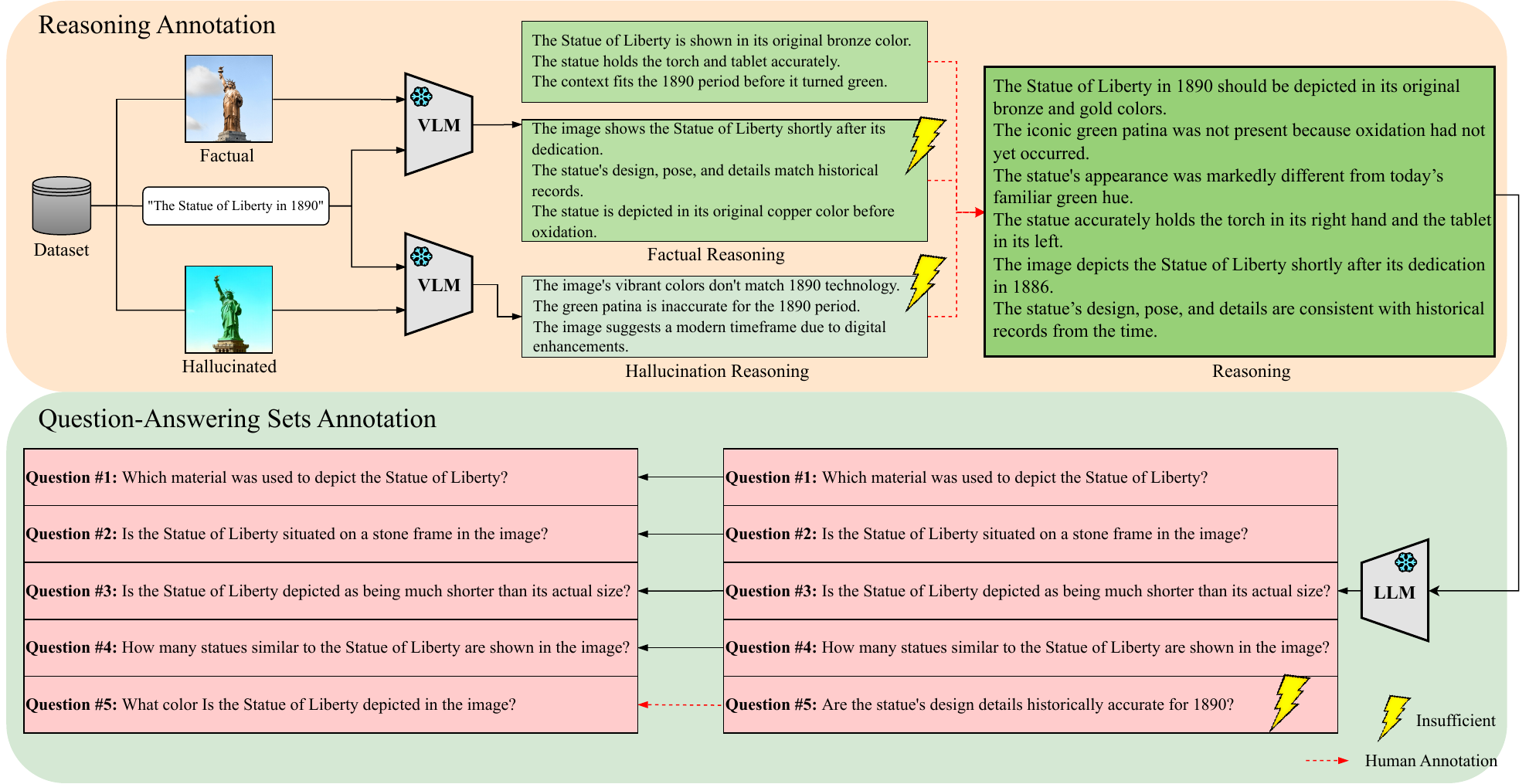} 
    \caption{The process of overcoming the rare insufficiency seen in GPT-4o through thorough human review, creating a meticulous benchmark and evaluation metric for assessing image hallucination. During enhancing our dataset, reasoning annotation is carried out. Reasoning is collected only when GPT-4o makes a correct prediction. The collected factual or hallucinated reasonings are reviewed by 10 participants and consolidated into a single reasoning. In the annotation stage when creating QA sets, GPT-4o generates QA sets based on this consolidated reasoning, and these are revised to reflect the most discriminative features that all 10 participants agreed upon.}
    \label{fig1}
\end{figure*}

\section{A. Challenges in Mitigating Image Hallucination by Text Prompt}

As depicted in Figure \ref{fig2}, image hallucination is not easily resolved by simply improving the text prompt. For example, when the prompt ``A molecule of water" is entered into DallE-3 \cite{dalle3}, the image obtained is shown in the first row of Figure \ref{fig2}. This is a hallucinated image for various reasons, such as the failure to accurately depict two hydrogen atoms and one oxygen atom. Therefore, we input this image and the prompt into GPT-4o to generate a detailed prompt designed to correct the image, much like how we obtained ``reasoning'' when enhancing the benchmark. We ask GPT-4o to suggest an alternative prompt that indicates how the original prompt should be modified to ensure the given image accurately reflects the prompt.
As a result, we get a long and detailed prompt, as shown in the second row of the figure. We then input this prompt back into DallE-3 to generate the image, and if the result is still a hallucinated image, we repeat the process. Even after several iterations of prompt refinement, it is evident that the text-to-image (TTI) model still struggles to generate an image that accurately reflects the factual information (such as the two hydrogen atoms, one oxygen atom, and the molecular structure).

\section{B. Eliminating Hallucination in GPT-4o: A Benchmark Refined by Human Review}
We propose I-HallA v1.0, a benchmark for evaluating image hallucination, utilizing GPT-4 Omni (GPT-4o) \cite{gpt4o}, one of the existing state-of-the-art vision-language models. 
This benchmark assesses external knowledge not included in the prompt by utilizing GPT-4o's extensive training data. It also evaluates visual semantics that are difficult to discern solely from prompts by leveraging GPT-4o's visual comprehension capabilities.
Although GPT-4o mostly generates accurate outputs, there are very rare instances where it fails to generate sufficient factual information or exhibits hallucination phenomena. Taking this into account, we have refined our benchmark and evaluation metrics through a rigorous human review process. Figure \ref{fig1} shows examples of the reasoning and QA sets before and after applying this human review process.

\section{C. Contrasting Image Hallucination with Common Sense}

\begin{figure}[!h]
    \centering
    \includegraphics[width=\columnwidth]{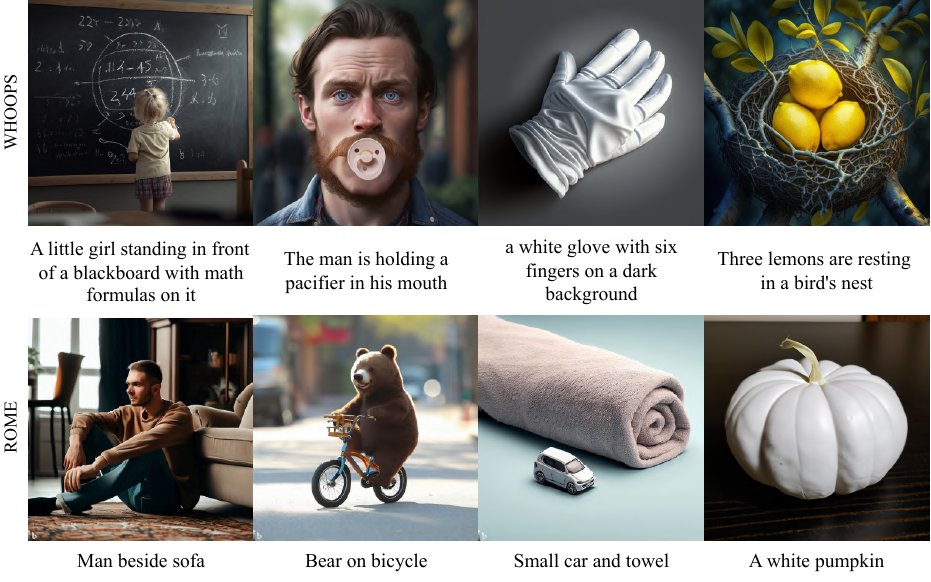} 
    \caption{Example images and corresponding prompts from existing research addressing common sense. These examples do not deal with image hallucination that contradicts factual information because, in many cases, they can sufficiently exist in the real world.}
    \label{fig3}
\end{figure}

In Figure \ref{fig3}, We introduce examples from WHOOPS \cite{whoops} and ROME \cite{rome}, existing benchmarks to evaluate the common sense reasoning abilities of vision-language models (VLMs) such as InstrcutBLIP \cite{instructBlip}. We demonstrate that the common sense addressed in previous studies differs from our proposed factual information and image hallucination.

The first row in Figure \ref{fig3} illustrates examples that violate common sense according to WHOOPS. However, a child solving a math problem or a man holding a pacifier in his mouth is not impossible or factually incorrect. Therefore, this is irrelevant to image hallucination.
The second row presents examples that violate common sense, according to ROME. Similarly, these examples cannot be considered impossible in the real world; thereby not dealing with image hallucination.

\begin{table*}[!h]
    \centering
    \small
    \begin{tabular}{ccc}
        \begin{minipage}[t]{0.28\textwidth}
            \centering
            \caption*{(a) InstructBLIP}
            \begin{tabular}{ccc}
                \toprule
                \multirow{2}{*}{Models} & \multicolumn{2}{c}{I-HallA Score} \\
                \cmidrule(lr){2-3}
                & Science & History \\
                \midrule
                SD v1.4 & 0.542 & 0.614  \\ 
                SD v1.5 & 0.556 & 0.606  \\
                SD v2.0 & 0.550 & 0.604 \\
                SD XL & 0.532 & 0.648 \\
                DallE-3 & 0.592 & 0.626 \\ \specialrule{0em}{1.5pt}{0pt}\cdashline{1-3}\specialrule{0em}{0pt}{1.5pt}
                Factual & \textbf{0.652} & \textbf{0.594} \\ 
                \bottomrule        
            \end{tabular}
        \end{minipage} &
        \begin{minipage}[t]{0.36\textwidth} 
            \centering
            \caption*{(b) GPT-4-turbo}
            \begin{tabular}{ccc}
                \toprule
                \multirow{2}{*}{Models} & \multicolumn{2}{c}{I-HallA Score} \\
                \cmidrule(lr){2-3}
                & Science & History \\
                \midrule
                SD v1.4 & 0.544 & 0.646  \\ 
                SD v1.5 & 0.516 & 0.62  \\
                SD v2.0 & 0.548 & 0.652 \\
                SD XL & 0.584 & 0.670 \\
                DallE-3 & 0.758 & 0.626 \\ \specialrule{0em}{1.5pt}{0pt}\cdashline{1-3}\specialrule{0em}{0pt}{1.5pt}
                Factual & \textbf{0.850} & \textbf{0.652} \\ 
                \bottomrule        
            \end{tabular}
        \end{minipage} &
        \begin{minipage}[t]{0.32\textwidth} 
            \centering
            \caption*{(c) GPT-4o-mini}
            \begin{tabular}{ccc}
                \toprule
                \multirow{2}{*}{Models} & \multicolumn{2}{c}{I-HallA Score} \\
                \cmidrule(lr){2-3}
                & Science & History \\
                \midrule
                SD v1.4 & 0.386 & 0.566  \\ 
                SD v1.5 & 0.40 & 0.556  \\
                SD v2.0 & 0.456 & 0.604 \\
                SD XL & 0.454 & 0.608 \\
                DallE-3 & 0.454 & 0.686 \\ \specialrule{0em}{1.5pt}{0pt}\cdashline{1-3}\specialrule{0em}{0pt}{1.5pt}
                Factual & \textbf{0.844} & \textbf{0.868} \\ 
                \bottomrule
            \end{tabular}
        \end{minipage} \\
    \end{tabular}
    \caption{I-HallA score table evaluating I-HallA v1.0 with three different VQA models. Compared to GPT-4o, which scored 0.856 in the science domain and 0.873 in the history domain for factual images, all three models demonstrate a lower ability to assess image hallucination. This is because, for factual samples, an I-HallA score closer to 1.0 is ideal, whereas the I-HallA scores of the three models are lower than that of GPT-4o.}
    \label{supp:t3}
\end{table*}

\begin{table*}[!htbp]
    \centering
    \begin{adjustbox}{width=\textwidth}
        \begin{tabular}{lp{0.5\textwidth}p{0.5\textwidth}}
            \toprule
            Aspects & Description & Example \\
            \midrule
            Prompt & A textual description of an image & A molecule of methane. \\
            Images & Either factual or hallucinated images. & (Factual image) \\
            Domain / Category & Which category it belongs to within education domain & Science / Biology \\
            Response & GPT-4o's results in distinguishing between factual and hallucinated images. & \{``Factual", ``Factual", ``Factual", ``Factual", ``Factual"\} \\
            Difficulty & Regarding how many times GPT-4o has answered correctly based on the responses: Easy: 4-5, medium: 2-3, hard: 0-1 & Easy \\
            Reasoning & Factual information should be considered to determine the hallucination in the given image. & The image accurately represents the structure of a methane molecule (CH4).
            
            - \textbf{Visual Evidence}: The illustration shows one central carbon atom labeled `C' with four hydrogen atoms labeled `H' surrounding it in a tetrahedral configuration. This geometry matches the known structure of methane, where the carbon atom forms single covalent bonds with four hydrogen atoms.
            
            - \textbf{Contextual Evidence}: Methane (CH4) is a well-documented simple hydrocarbon and the simplest alkane. Its molecular structure, consisting of one carbon atom bonded to four hydrogen atoms, is commonly represented in scientific contexts. \\
            VQA & Question-answering sets that evaluate the image hallucination \& factual information of a given image. & \textbf{Question}: What is depicted at the molecule's center in the image?
            
            \textbf{Choices}:
            
            A) Carbon atom
            
            B) Hydrogen atom
            
            C) Oxygen atom
            
            D) Nitrogen atom
            
            E) None of the above\\
            CoI & The Composition of Interest (CoI) that each question is most relevant to. & Existence \\
            \method Score & A score reflecting how much factual information the given image represents. & 1.0 \\
            \bottomrule        
        \end{tabular}
    \end{adjustbox}
    \caption{Illustrations of the sequence-to-sequence formatting for each aspect (images omitted but are part of the aspects).}
    \label{tab:t1}
\end{table*}

\begin{table*}[!htb]
    \centering
    \begin{adjustbox}{width=\textwidth, height=\textheight, keepaspectratio}
        \begin{tabular}{lp{0.5\textwidth}p{0.5\textwidth}}
            \toprule
            Composition & Definition & VQA \\
            \midrule
            \textbf{Color} & The visual appearance of objects in terms of hue, saturation, and brightness, distinguishing one object from another. & \textbf{Question}: What are the predominant colors in the flag visible in the image?
            
            \textbf{Choices}:
            
            A) Green and yellow
            
            B) Red, white, and blue
            
            C) Black and white
            
            D) Orange and purple 
            
            E) None of the above\\ 
            \textbf{Counting} & Determining the number of objects or entities in a given image or scenario. & \textbf{Question}: How many hydrogen atoms are connected to the central atom in the image?
            
            \textbf{Choices}:
            
            A) Two
            
            B) Three
            
            C) Four
            
            D) Five
            
            E) None of the above\\ 
            \textbf{Existence} & Confirmation that basic entities within an image, such as a person, animal, food, items, vehicles, or text symbols (e.g., “A”, “1+1”) are correctly present and accurately depicted. & \textbf{Question}: What is depicted at the center of the molecule in the image?
            
            \textbf{Choices}:
            
            A) Carbon atom
            
            B) Hydrogen atom
            
            C) Oxygen atom
            
            D) Nitrogen atom 
            
            E) None of the above\\ 
            \textbf{Others} & Any other characteristics not covered by specific categories, such as abstract properties. & \textbf{Question}: When do you think the artifact was created in the image?
            
            \textbf{Choices}:

            A) Ancient
            
            B) Modern
            
            C) Medieval
            
            D) Ice age 
            
            E) None of the above\\
            \textbf{Posture} & The orientation or positioning of a person or object, indicating actions, movements, or stances. & \textbf{Question}: What is the posture of the statue in the image?
            
            \textbf{Choices}:
            
            A) Sitting
            
            B) Standing 
            
            C) Lying down
            
            D) Running 
            
            E) None of the above\\
            \textbf{Relation} & Connections between entities, including spatial arrangements (e.g., on top, inside) and part-whole connections (e.g., body parts, clothing). & \textbf{Question}: How are the hydrogen atoms positioned relative to the central carbon atom in the image?
            
            \textbf{Choices}:
            
            A) Opposite sides
            
            B) Equidistant
            
            C) Adjacent
            
            D) Randomly 
            
            E) None of the above\\
            \textbf{Scene} & Backgrounds or settings of an image, such as weather, location, or environmental context. & \textbf{Question}: Where does the image appear to be set?
            
            \textbf{Choices}: 
            
            A) A forest
            
            B) A hockey rink
            
            C) A beach
            
            D) A classroom
            
            E) None of the above\\
            \textbf{Shape} & The geometric form or outline of objects, helping to distinguish between different objects or entities. & \textbf{Question}: Which of the following shapes is prominent in the image?
            
            \textbf{Choices}:
            
            A) Circular shapes of bubbles
            
            B) Square buildings
            
            C) Flames with irregular shapes
            
            D) Triangular tents
            
            E) None of the above\\
            \textbf{Size} & The dimensions or scale of objects relative to others within the image, such as large, small, tall, or wide. & \textbf{Question}: What is the size of the container holding the water described in the image?
            
            \textbf{Choices}:
            
            A) Larger than human
            
            B) Small beaker
            
            C) Medium-sized pot
            
            D) A 1-meter water tank
            
            E) None of the above\\
            \bottomrule        
        \end{tabular}
    \end{adjustbox}
    \caption{Detailed description of Compositions of Interest (CoIs). Each CoI is defined along with a corresponding QA set example.}
    \label{suppl:t2}
\end{table*}

\begin{figure}[!ht]
    \centering
    \includegraphics[width=0.95\columnwidth, height=\textheight, keepaspectratio]{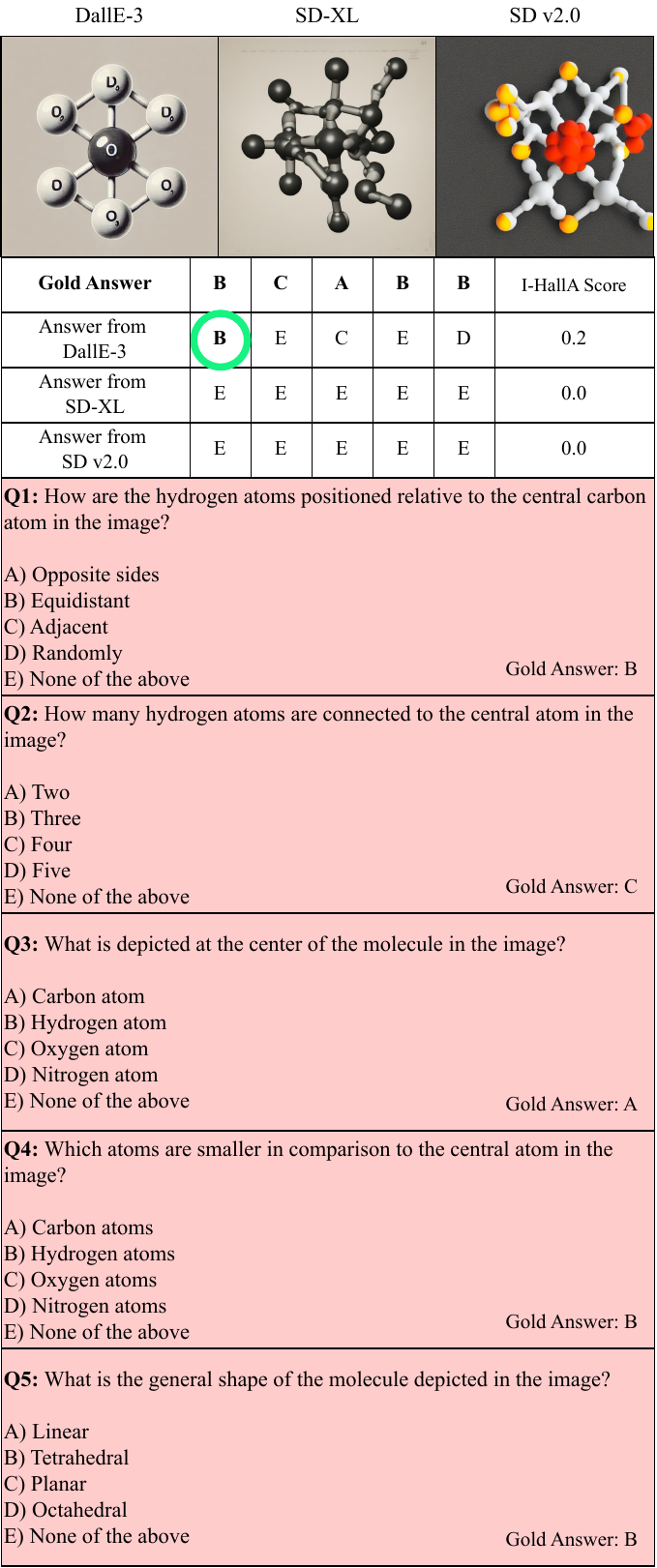} 
    \caption{The process of calculating I-HallA scores for given images. The VQA model answers questions based on the image, and these answers are compared to the gold answers corresponding to the factual information to measure the extent of factual content in the image.  A higher score indicates a greater reflection of factual information and less image hallucination.}
    \label{fig5}
\end{figure}

\section{D. I-HallA vs. Existing Metrics for Text-to-Image Models: A Detailed Comparison}
\label{supp:existing_metrics_compare}

Metrics such as TIFA \cite{TIFA}, VQ\textsuperscript{2} \cite{vq2_seetrue}, and Davidson Graph Scene \cite{dsg}, which evaluate the alignment of TTI models, are limited by being generated solely from the given text prompt. For example, when applying TIFA, the most representative TTI metric, to the prompt ``A molecule of methane," TIFA first uses GPT-3 \cite{gpt3} to extract elements from the prompt: \emph{molecule, methane}

Next, it generates two questions for each element. The first question should receive a ``yes'' answer if the image is correctly generated, and the second question is one where the element itself should be the answer: \emph{\{``Is there a molecule in the image?'', Choices: ``Yes,'' ``No''\}, \{``What is this?", Choices: ``atom,'' ``molecule,'' ``ion,'' ``compound"\}}.

The TIFA score obtained from these questions tends to be high for both factual and hallucinated images, making it inadequate for evaluating image hallucination. Like TIFA, other metrics also generate QA sets based on the prompt through an analogous process. Consequently, as described in the introduction and methodology, these metrics cannot evaluate factual information beyond the prompt and cannot properly assess realistic visual semantics due to the polysemy of the prompt.

\section{E. Ablation Study with Various Visual Question Answering (VQA) Models}

In this section, we evaluate five TTI models on image hallucination using various visual question-answering (VQA) models. The TTI models include Stable Diffusion v1.4 \cite{stablediff}, Stable Diffusion v1.5 \cite{stablediff}, Stable Diffusion v2.0, Stable Diffusion XL-Base v1.0, and DallE-3, while the VQA models include InstructBLIP \cite{instructBlip}, GPT-4-turbo \cite{gpt4turbo}, and GPT-4o-mini \cite{gpt4omini}.

In Table \ref{supp:t3}-(a), \textbf{InstructBLIP} shows little to no difference in \method scores between factual and hallucinated images, proving that the VQA model cannot distinguish between them. Notably, the model fails to recognize factual images in both the science and history domains, with I-HallA scores of 0.652 and 0.594, respectively. Given that an ideal I-HallA score for factual images would be 1.0, these scores are significantly lower than the 0.856 and 0.873 recorded by GPT-4o for science and history, respectively, which can be interpreted as a limitation of the VQA model.

In Table \ref{supp:t3}-(b), \textbf{GPT-4-turbo} shows slightly better results than InstructBLIP; however, it still fails to correctly assess factual images, particularly in the history domain. In the science domain, there are instances where hallucinated images are incorrectly judged as factual, such as with DallE-3, which received an I-HallA score of 0.850.

In Table \ref{supp:t3}-(c), \textbf{GPT-4o-mini} is comparable to GPT-4o, but it still cannot accurately identify factual images as well as GPT-4o, recording lower scores in both the science and history domains.
From these results, we conclude that GPT-4o is the most suitable VQA model for evaluating various TTI models on image hallucination. Additionally, the effective use of our metric requires a highly advanced VQA model, such as GPT-4o. This underscores the difficulty of accurately assessing factual information and evaluating image hallucination within images. This challenge highlights the pioneering nature and novelty of our research.

\begin{figure*}[htp]
    \centering
    \includegraphics[width=1.0\textwidth]{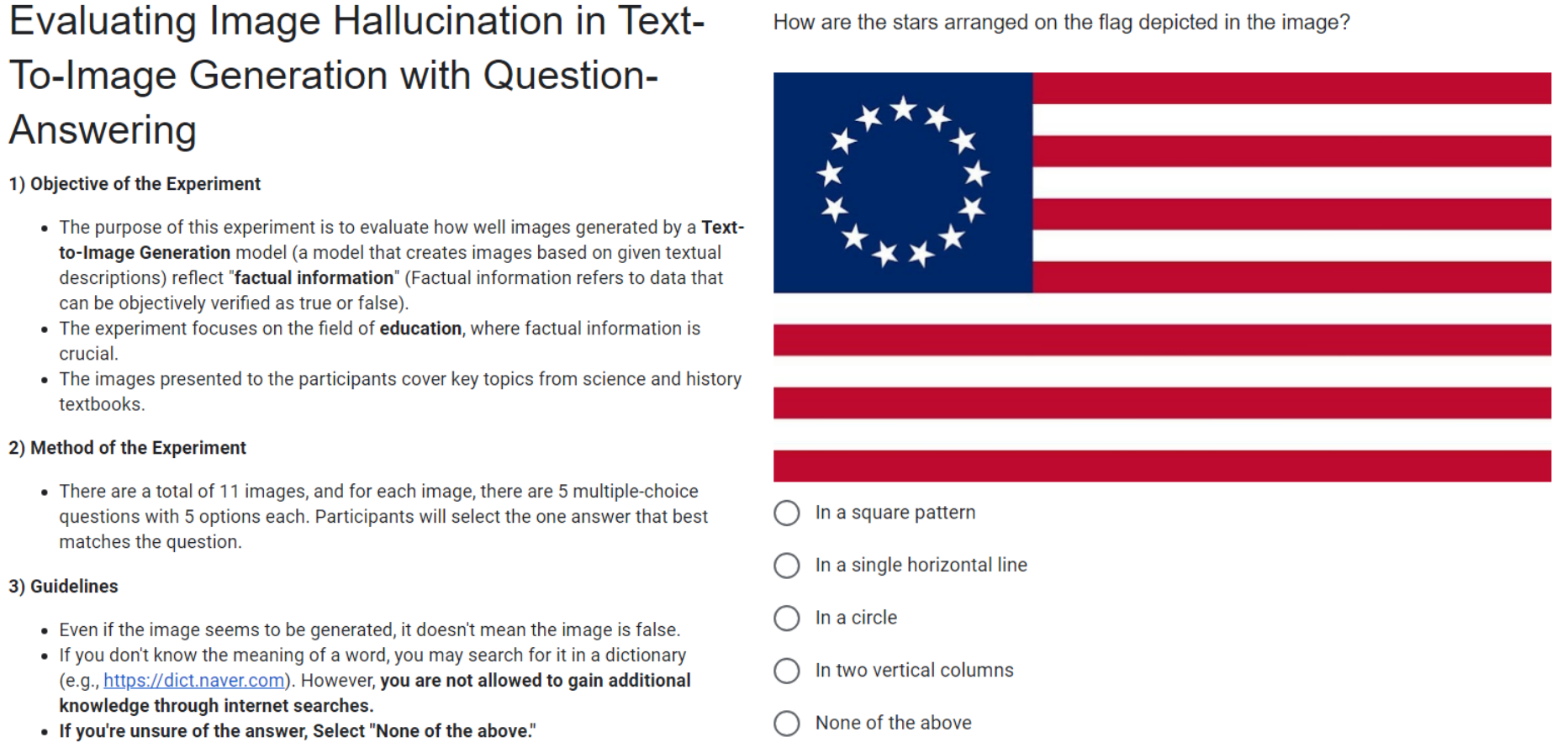} 
    \caption{The instructions and example question shown to participants during the human evaluation process. }
    \label{fig6}
\end{figure*}

\section{F. I-HallA Score Computation}

Figure \ref{fig5} shows hallucinated images generated by three TTI models (DallE-3, Stable Diffusion XL, Stable Diffusion v2.0) for the prompt ``A molecule of methane.'' It also includes the corresponding I-HallA QA sets, the answers generated by GPT-4 after viewing the images and questions, and the resulting I-HallA scores.  Each question is designed to assess elements such as the number of atoms and the molecular structure, thereby evaluating the extent of hallucination in the given prompt and generated image.

For the hallucinated image generated by DallE-3, the equidistant placement of atoms around the central atom allows it to correctly answer the first question, matching the gold answer. However, the remaining questions reveal that the image fails to accurately depict factual information about the methane molecule, resulting in answers different from the gold answers. Consequently, with one correct answer out of five questions, it records an I-HallA score of 0.2. The other TTI models exhibit even more severe hallucinations, failing to correctly answer any of the five questions, thus scoring 0.

\begin{figure*}[!h]
    \centering
    \includegraphics[width=1.0\textwidth, height=\textheight, keepaspectratio]{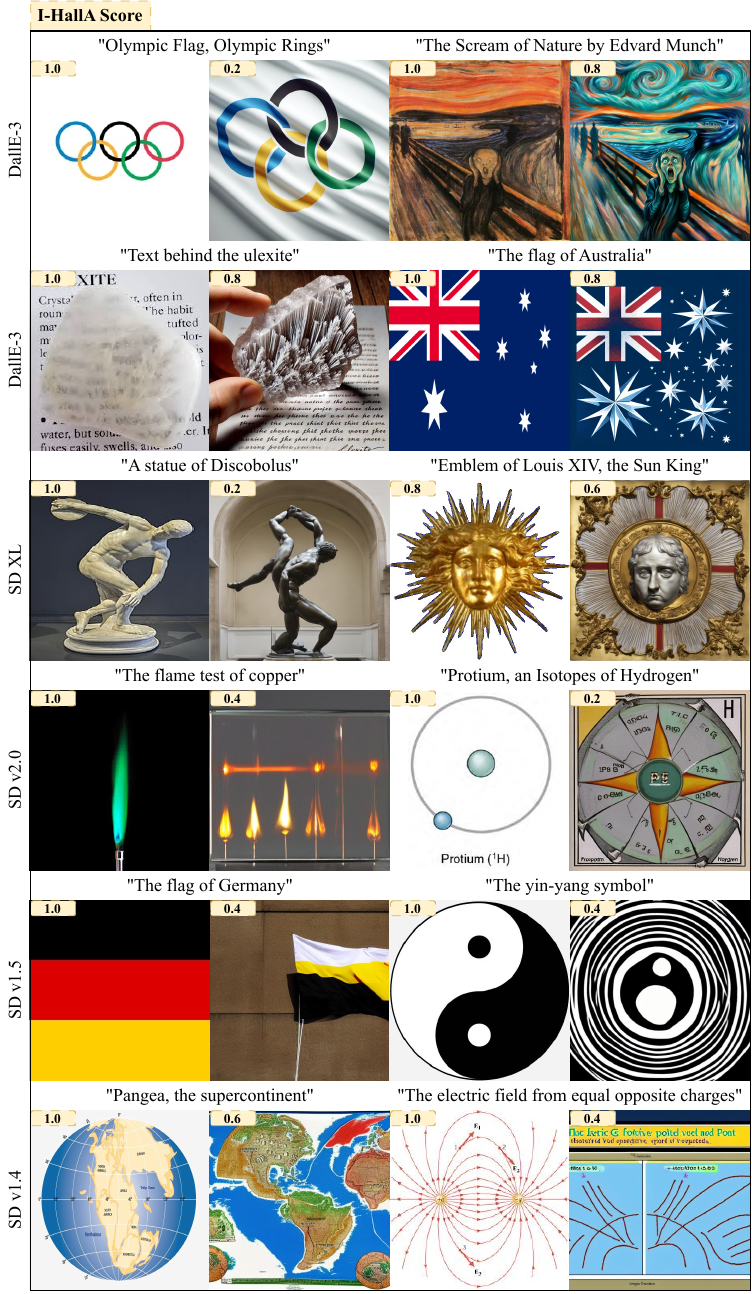} 
    \caption{The image examples of I-HallA v1.0 span across various prompts. For each prompt, the left image is factual, while the right image is a hallucinated image generated by different text-to-image models. The I-HallA score for each image is displayed in the top left corner.}
    \label{fig10}
\end{figure*}

\begin{figure*}[!h]
    \centering
    \includegraphics[width=1.0\textwidth, height=\textheight, keepaspectratio]{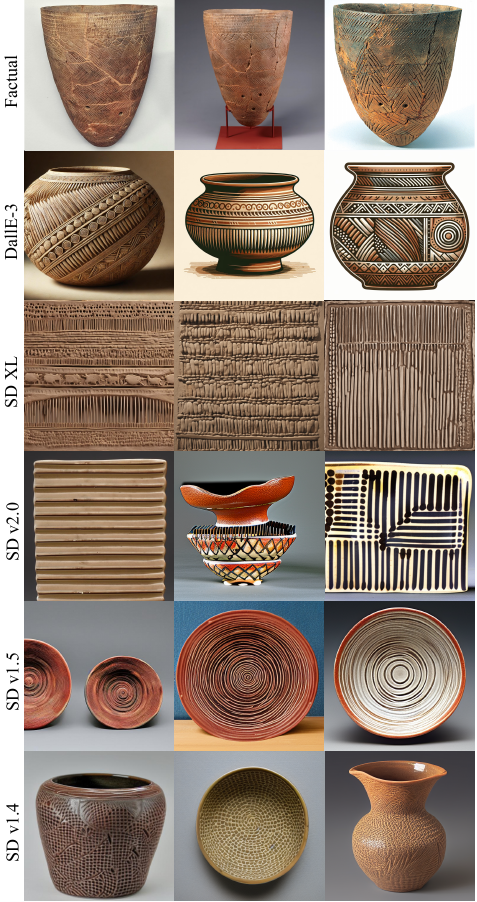} 
    \caption{Examples of different visual semantics generated by five TTI models for the prompt ``Comb-pattern pottery BCE 8000.'' While Stable Diffusion v2.0, Stable Diffusion XL, and DallE-3 reflect the comb-pattern, they produce pottery that is not factually accurate.}
    \label{fig7}
\end{figure*}

\section{G. Details for Human Evaluation}
For human evaluation, we recruit 53 participants, including 32 males and 21 females. The participants have an age distribution of 21 individuals aged 18-24, 31 aged 25-34, and 1 aged 35-44. Their educational backgrounds are distributed as follows: 12 are currently enrolled in university, 16 have graduated from university, 21 are enrolled in graduate school, and 4 have graduated from graduate school.

The test set for analysis in the human evaluation consists of 10 images and 50 QA sets.
We additionally insert one dummy image with 5 corresponding QA sets at the beginning, which we exclude from the final statistical analysis.
We randomly select 4 factual images and 3 hallucinated images from each of DallE-3 and SD-XL, ensuring that prompts do not overlap. The instructions and example questions provided to the participants are shown in Figure \ref{fig6}. As in the I-HallA score experiments using the VQA model, we ask five questions corresponding to each prompt. To ensure participants focus solely on the visual elements of the images, we do not reveal the prompts.

\section{H. I-HallA v1.0: Exploring the Benchmark Through Detailed Examples}
Figure \ref{fig10}, Table \ref{tab:t1}, and Table \ref{suppl:t2} show examples of the I-HallA v1.0 image dataset, examples of the overall aspects, and examples of Compositions of Interest (CoI), respectively.
Figure \ref{fig10} presents 12 randomly selected prompts along with one corresponding factual image and one hallucinated image, accompanied by the I-HallA score. DallE-3 introduces 4 images, while the remaining Stable Diffusion models introduce 2 images each. 
In Table \ref{tab:t1} various aspects that appear throughout the entire process of this paper, from the prompt to the calculation of the I-HallA score, are explained with examples.
Table \ref{suppl:t2} describes the concepts of each composition, providing examples of corresponding QA sets.

\section{I. Qualitative Examples of Polysemy in Text Prompts}
Figure \ref{fig7} introduces detailed examples of the various visual semantics that can be generated due to the polysemy of text prompts. The first row shows factual images for the prompt "Comb-pattern pottery BCE 8000." Commonly, these images feature conical-shaped brown pottery with diagonal comb patterns. However, in the images generated by Stable Diffusion v1.4 and v1.5 \cite{stablediff}, the comb pattern on the pottery is not accurately depicted. While Stable Diffusion v2.0, Stable Diffusion XL, and DallE-3 do generate pottery with comb patterns, the forms and patterns are significantly different from factual images. This demonstrates that although TTI models can generate images that reflect the text prompt (comb-pattern), they fail to capture the factual visual semantics due to the polysemy of the text. Additionally, this implies that existing evaluation methods, which rely on prompts, cannot properly assess the false comb-pattern pottery present in the images in the 2nd to 4th rows of Figure \ref{fig7}.

\FloatBarrier
\section{J. Prompt}
For demonstration purposes, we present a portion of the prompt for our three-stage pipeline, which utilizes multiple GPT-4o-based agents. We introduce three representative GPT-4o-based agents for reasoning generation, question generation, and evaluation within the \method. We provide each agent with a purpose-specific prompt, which includes instructions and several in-context examples. The examples are taken from outside the \benchmark to ensure fairness. We omit part of the first phase of our pipeline, as it is conducted exclusively by human annotators. The complete prompt will be released alongside our codes.
Each stage includes a final human review of the results as the last step.

\subsection{Reasoning Agent}
The reasoning agent aligns with the second phase of our three-stage pipeline, which aims to generate curated reasoning for a given image and its corresponding prompt.
Using the provided prompt, GPT-4o generates a response and reasoning, producing a total of 5 results per image as we repeat the task five times.
The reasoning includes both visual and contextual evidence from the image.
We discard those whose responses do not correctly answer whether the image is factual or hallucinated, as we expect their reasonings to be unreliable.

\begin{lstlisting}
Your Role: Excellent Image Distinguisher

Objective: Given the input image and caption, Distinguish whether the given image is normal (reasonable and factual), or if it is weird (incorrect, distorted, and contrary to common sense).
More specifically, if the truth is important such as science and history, you might need to say "weird" for the given image is different from the fact; otherwise "normal".

Your reasoning must be as detailed as it covers the explanations about almost every component within the image.
When determining if an image is normal or weird, do not judge it solely based on whether it has a virtual or cartoon-like style.

You should follow the response format below:

[Response Format]
- Response: <Normal or Weird>
- Reasoning: <Your reasoning about why you have made this decision>
  - **Visual Evidence**: <Your reasoning about why you made this decision solely based on the visual appearance>
  - **Contextual Evidence**: <Your reasoning about why you made this decision solely based on the context>

When evaluating an image, use very detailed aspects of the image and compare them with reality as the basis for your judgment.
For example, consider the number of objects in the image, the shapes of structures, the number of floors in buildings, and other similar information. Compare these details with real-life facts to distinguish and identify the image accurately.

We provide you with a few-shot examples to help you better handle your task.

For each example below, an input image is given and paired with three parameters: the caption for the normal image, the response, and the reasoning.
Considering these few-shot examples, we expect you to successfully achieve your current task.

[Example]
### Example {example_number}
- Caption: {example_caption}
- Response: {example_response}
- Reasoning: {example_reasoning}
- Image:

Your Current Task:
Given the input image and its corresponding caption, distinguish whether it is weird or normal. Ensure adherence to the response format provided above.

Caption: {caption}
Image:

\end{lstlisting}

\subsection{Questiong-Answering Agent}
The Question-Answering agent is involved in the third phase of our pipeline, returning five pairs of questions and answers per image. As mentioned in the paper, this agent provides a question-choices-answer set and the corresponding composition of interest.

\begin{lstlisting}
Your Role: Questions and Answers Generator

Objective:
Given a sample that includes a prompt and textual reasoning that explains whether the given image is normal (reasonable and factual) or weird (hallucination, incorrect, distorted, and contrary to common sense), generate multi-choice questions that verify if the image is normal or weird.
When a correctly or incorrectly generated image is provided for a given prompt, these QA(Question and Answers) sets are designed to accurately assess and distinguish the factual information regarding the image.
The last choice should be "None of the above."

You should also match each QA with the most related compositions of interest (CoI).
The table of CoIs includes existence, size, color, shape, posture, relation, scene, and counting.
If the generated QA does not match any of the CoIs listed above, select "others."

Constraints:
- If the correct answer to a question in the QA set is given for a given image, the image should contain factual information. In other words, an incorrectly generated image should not have the correct answer.
- In reasoning, exclude context evidence (i.e., "The Scream of Nature" falls under which art movement?) that cannot be visually recognized. In other words, you should only generate QA sets based on visual evidence; only referring to context evidence for better understanding.
- Avoid including explicit expressions (i.e., "The Scream of Nature", "Edvard Munch", etc.) in your QA. Rather, replace them with general expressions like "the image".

You should follow the response format below:

[Response Format]
Set 1:
Compositions of Interest:
Question:
Choices:
A)
B)
C)
D)
E) None of the above
Answer:

Set 2:
Compositions of Interest:
Question:
Choices:
A)
B)
C)
D)
E) None of the above
Answer:

Set 3:
Compositions of Interest:
Question:
Choices:
A)
B)
C)
D)
E) None of the above
- Answer:

Set 4:
Compositions of Interest:
Question:
Choices:
A)
B)
C)
D)
E) None of the above
Answer:

Set 5:
Compositions of Interest:
Question:
Choices:
A)
B)
C)
D)
E) None of the above
Answer:

Your Current Task:
Given the reasoning for an image, generate five possible QA sets. Ensure adherence to the response format provided above.

Caption: {caption}
Reasoning: {reasoning}
\end{lstlisting}

\subsection{Evaluation Agent}
The Evaluation agent represents the final stage of our pipeline, scoring \method for a given image. Specifically, when GPT-4o determines that none of the choices correctly answer the question, it can return ``E) None of the above".
We then calculate the \method score for the image by comparing the predicted answers from the five questions with the ground-truth answers. For example, if 2 out of 5 questions are answered correctly, we assign a \method score of 0.4 for that image.

\begin{lstlisting}
You are an agent who answers questions based on the given image.
Here, the given image could be wrong.

So, your job is to either choose the best answer choice or output "None of the above" for ambiguous questions.

Your answer must choose one of the given choices to your best knowledge.
It must be a single character that indicates its answer (i.e., "A", "B", "C", "D", "E").
As shown below [Example], your answer should not include texts other than a single character ("A", "B", "C", "D", and "E").

[Example]
Question: How many continents are shown as connected in the image?
Choices:
A) Korean Peninsula
B) One large supercontinent
C) Two separate continents
D) Three continents joined together
E) None of the above
Answer: B

Question: How many atoms of the molecule in the image?
Choices:
A) 1
B) 2
C) 3
D) 4
E) None of the above
Answer: E

DO NOT use your external knowledge obtained by pre-training large-scale data.
You must base your judgment on the image input's visual evidence.

You should answer the following question:

[Question]
Question: {question}
Choices: {choices}

[Response Format]
Answer:
\end{lstlisting}

\end{document}